\pdfoutput=1

\documentclass[11pt]{article}

\usepackage{acl}

\usepackage{times}
\usepackage{latexsym}

\usepackage[T1]{fontenc}

\usepackage[utf8]{inputenc}

\usepackage{microtype}

\usepackage{inconsolata}

\usepackage{babel} 
\usepackage{graphicx}
\usepackage[subrefformat=parens]{subcaption}
\usepackage{amsmath}

\usepackage{multirow}
\usepackage{tabularx}
\usepackage{makecell}
\usepackage{booktabs}

\usepackage[ruled]{algorithm2e}
\SetKwComment{Comment}{/* }{ */}
\SetKw{Break}{break}

\usepackage{hyperref}
\usepackage[nameinlink]{cleveref}

\newcommand{\qblank}[1]{%
    \textbf{\_\_#1\_\_}
}

\newcommand{\footremember}[2]{%
    \footnote{#2}
    \newcounter{#1}
    \setcounter{#1}{\value{footnote}}%
}
\newcommand{\footrecall}[1]{%
    \footnotemark[\value{#1}]%
}

\title{DisGeM: Distractor Generation for Multiple Choice Questions\\with Span Masking}

\author{
    Devrim Çavuşoğlu \thanks{Equal contribution.} \footremember{obss}{OBSS AI} \footremember{metu}{Middle East Technical University}%
    \and Seçil Şen \footnotemark[1] \footrecall{obss} \footremember{boun}{Bogazici University} %
    \and Ulaş Sert \footrecall{obss} \\ %
    \footrecall{obss} \; OBSS AI \\
    \footrecall{metu} \; Middle East Technical University \\
    \footrecall{boun} \; Bogazici University \\
    first-name.last-name@obss.tech
    }

\begin{document}
\maketitle
\begin{abstract}
Recent advancements in Natural Language Processing (NLP) have impacted numerous sub-fields such as natural language generation, natural language inference, question answering, and more. However, in the field of question generation, the creation of distractors for multiple-choice questions (MCQ) remains a challenging task. In this work, we present a simple, generic framework for distractor generation using readily available Pre-trained Language Models (PLMs). Unlike previous methods, our framework relies solely on pre-trained language models and does not require additional training on specific datasets. Building upon previous research, we introduce a two-stage framework consisting of candidate generation and candidate selection. Our proposed distractor generation framework outperforms previous methods without the need for training or fine-tuning. Human evaluations confirm that our approach produces more effective and engaging distractors. The related codebase is publicly available at \href{https://github.com/obss/disgem}{https://github.com/obss/disgem}.
\end{abstract}

\section{Introduction}
\label{sec:intro}

Multiple-choice cloze tests are a prevalent form of assessment that not only evaluates a student's reading comprehension but also challenges their ability to deduce the most fitting option from a set of alternatives. Rooted in the concept of the traditional cloze test, where specific words are omitted from a passage and students are required to fill in the blanks with appropriate terms, the multiple-choice variant enhances the testing methodology by presenting a selection of potential options for each blank. This approach adds an element of complexity, requiring students to not only comprehend the context but also discern the most contextually appropriate answer among the provided choices. Central to the construction of multiple-choice cloze tests are the distractors – options deliberately crafted to divert students away from the correct answer. The creation of these distractors involves a careful balance of linguistic nuances while also appearing plausible enough to challenge the analytical skills of test-takers. In the realm of education, these tests serve as valuable tools for educators to gauge students' reading comprehension, critical thinking, and inference abilities, offering a holistic assessment of their grasp on the subject matter. We provided an example of a multiple-choice cloze style question in \Cref{tab:cloth-ex-gold} from the CLOTH dataset \cite{xie-etal-2018-large}.

\begin{table}
\centering
\small
    \begin{tabularx}{\linewidth}{c|X}
        \toprule
        \textbf{Stem} &   If you are grateful, you naturally \qblank{\_} yourself up to receive all kinds of blessings and good things in life.  \\
        \midrule
        \textbf{Options} & \makecell[l]{\textbf{A. open} \\  B. make \\ C. stand \\ D. take}  \hfill \makecell[l]{$\rightarrow$ Answer \\ $\rightarrow$ Distractor \\ $\rightarrow$ Distractor \\ $\rightarrow$ Distractor} \hfill\null \\
    
        \bottomrule
    \end{tabularx}
    \caption{A Cloze Test Example from CLOTH Dataset: The challenge of multiple-choice cloze test generation pertains to the creation of both plausible and reliable distractors.}
    \label{tab:cloth-ex-gold}
\end{table}


Convincing distractor generation for short-form extractive multiple-choice questions (MCQ) in NLP remains an active research area, offering potential applications in educational assessments and question-generation tasks \cite{agarwal-mannem-2011-automatic}. This paper presents a novel approach to generating distractors by leveraging the capabilities of Pre-trained Language Models (PLMs), specifically, those based on the Transformer architecture \cite{vaswani2017transformer}. Our proposed technique aims to generate distractors that closely resemble correct answers while maintaining semantic dissimilarity, all while utilizing publicly available pre-existing models like BERT \cite{devlin-etal-2019-bert} and RoBERTa \cite{liu2019roberta}.

Unlike prior approaches that rely on parts-of-speech, recurrence, WordNet, or semantic analysis \cite{madri2023comprehensive}, our approach utilizes Transformer-based PLMs to generate automated, diverse, and plausible distractors for short-form extractive MCQs without the need for additional model training. We make use of PLMs trained with the Masked Language Modeling (MLM) paradigm and Natural Language Inference (NLI), tasks for which extensive pre-trained models exist \cite{devlin-etal-2019-bert,liu2019roberta,dagan2005pascal}. MLM training allows these models to comprehend contextual information and generate coherent text, while NLI helps in maintaining semantic dissimilarity, creating distractors that closely mirror correct answers but differ in meaning.

MLM naturally aligns with distractor generation, as it draws inspiration from the Cloze task \cite{taylor1953cloze}. The Cloze task involves completing a text with omitted words based on contextual understanding, paralleling our use of MLM for distractor generation. By aligning these two concepts, we leverage MLM's innate strength in contextual completion to generate more relevant and coherent distractors.

Furthermore, we tackle a challenge presented by Transformer-based PLMs: their limitation in generating content within pre-existing text. While these models excel at MLM tasks, they require a predetermined token count for the masked region, lacking flexibility in dynamically determining the length of the generated text. This limitation poses a challenge since excessively long or short distractors may adversely impact their plausibility. To overcome this, we propose a system that allows for the generation of distractors of variable lengths, ensuring they remain within an acceptable range.

\begin{figure}
\centering
\includegraphics[width=0.95\linewidth]{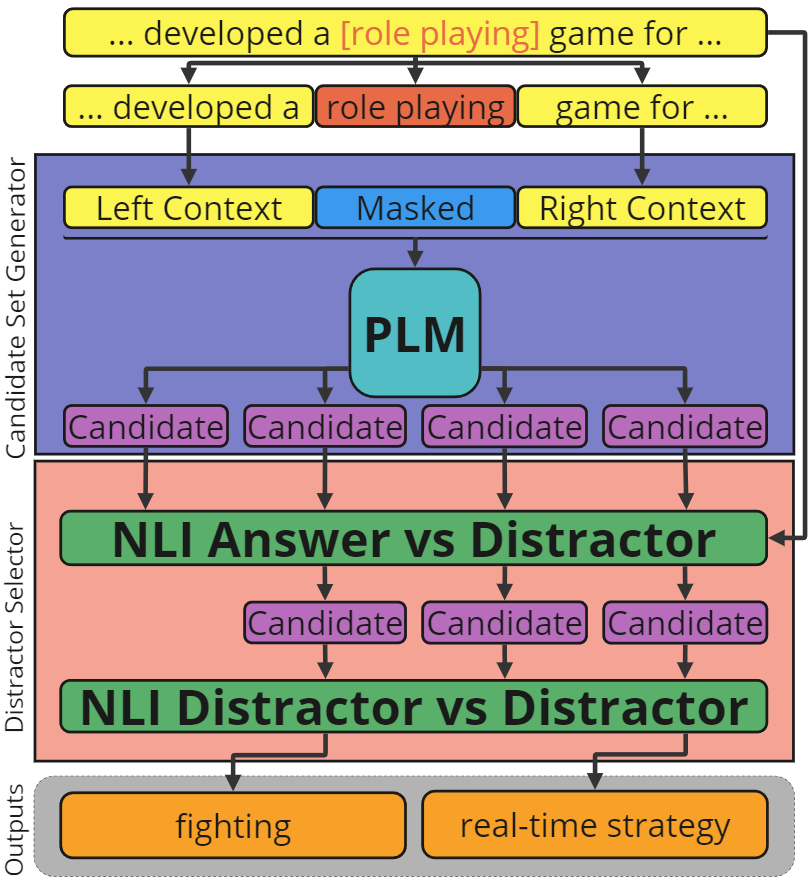}
\caption{The overall architecture of DisGeM. Pre-trained Language Model generates candidates, which are then filtered with two NLI models to ensure consistency among the correct answer and distractors.}
\label{fig:disgem}
\end{figure}

This study builds on the CDGP framework \cite{chiang-etal-2022-cdgp} by introducing a new n-gram token generation procedure and refining the distractor selection method, outlined in \Cref{fig:disgem}. The key contributions of our work are threefold. Firstly, we demonstrate that our system can produce a diverse set of distractors that exhibit similar characteristics to the correct answers, enhancing the sophistication of short-form extractive MCQs. Secondly, we emphasize that our method requires no specific training, making it easily applicable to various domains and languages without the need for extensive data collection. Thirdly, our framework offers a structured approach that can be readily extended or modified to accommodate different requirements, promoting adaptability and scalability. While our performance on automated metrics is comparable to the previous work, from the human evaluations we observe a clear preference toward our framework.


\section{Related Work}
\label{sec:rel-work}

Early NLP research focused on automatic question generation, with particular attention to Multiple-Choice Questions (MCQs), addressing the challenge of distractor generation, i.e. generating plausible distractors alongside correct answers \cite{mcqsurvey}. Traditional methods, including semantic analysis and ontologies, were initially employed for distractor generation \cite{kumar-2023-distractor, ha-yaneva-2018-automatic, Faizan2018AutomaticGO}. Recent approaches leverage ontologies to select distractors based on semantic relationships \cite{Ren_Zhu_2021, liang-etal-2018-distractor}.

Advancements in deep learning introduced end-to-end frameworks such as Bi-LSTM and sequence-to-sequence models for distractor generation \cite{qiu-etal-2020-automatic}. Transformer models further improved distractor generation, with approaches like round-trip neural machine translation \cite{panda-etal-2022-automatic} and one encoder, three decoders architecture \cite{learningtodistract}. Transformer fine-tuning strategies have also been applied \cite{chiang-etal-2022-cdgp, offerijns2020better, chung-etal-2020-bert}.

Two noteworthy frameworks emerged, both featuring a candidate set generator and a distractor selector \cite{Ren_Zhu_2021, chiang-etal-2022-cdgp}. \citet{Ren_Zhu_2021} integrates knowledge bases, ensuring semantic and grammatical relatedness, and uses a learning-to-rank model for distractor selection. Building on this, \citet{chiang-etal-2022-cdgp} proposes a pre-trained language model-based distractor generator that outperforms the former. However, fine-tuning limitations and computational costs are noted. In contrast, our model harnesses pre-trained masked language models without requiring fine-tuning, offering a simpler and more practical solution.

\section{Methodology}
\label{sec:method}

Our approach, building upon the CDGP framework \cite{chiang-etal-2022-cdgp}, is a dual-stage procedure: first, a candidate set generator (CSG) creates potential options, followed by a distractor selector (DS) that finalizes the distractor set. See \Cref{fig:disgem} for our framework's structure. Notably, instead of requiring a training regimen, our methodology leverages a pre-trained language model to produce candidates. In the second stage, these candidates undergo a meticulous two-step elimination process to select the final distractors. For a detailed view of the distractor generation pipeline and the algorithm, we refer readers to \Cref{apx:pipe}.

\subsection{Candidate Set Generator (CSG)}
\label{ssec:csg}

The first phase of our framework, the generation phase, generates distractor candidates using the source context $S$ and the answer $a$ and $a = [a_1 || a_2 || ... || a_r]$ with $a_i$ being the $i^{th}$ token string of the answer string $a$, and $[\cdot||\cdot]$ denotes concatenation of strings. In this framework, the answer is assumed to be a span/substring of the given context, i.e. $a \in S$. To generate candidates we use any PLM trained with an MLM objective \cite{devlin-etal-2019-bert, liu2019roberta, lan2019albert, he2020deberta}. With a chosen model, this phase involves two steps: first, masking the tokens of the answer $a$, and second, generating multiple candidates from these inputs. This phase corresponds to the distractor generation stage represented by the blue area in \Cref{fig:disgem}. Although similar to the CSG phase in the foundational CDGP framework \cite{chiang-etal-2022-cdgp}, DisGeM's CSG exhibits two noteworthy advantages: \textbf{(i)} it doesn't require fine-tuning, but allows it as an option for tailored applications (e.g., domain adaptation), and \textbf{(ii)} it can generate multi-word candidates, in contrast to CDGP's capacity limited to single-word candidates.

Unlike CDGP, our CSG is capable of generating multi-word/token candidates. However, a challenge arises when attempting to predict all mask tokens simultaneously. The model might produce suboptimal results because it lacks awareness of the best predictions for the nearby masked tokens. To address this, we patch an auto-regressive generation strategy and incorporate decoding techniques such as left-to-right (L2R) and right-to-left (R2L) decoding \cite{watanabe-sumita-2002-bidirectional}. For example, with L2R decoding, tokens are generated in a left-to-right sequence one-by-one, so that every other token would be aware of the surrounding context and conditioned on the previous generations. 

We employ a beam search \cite{graves2012sequence} alike algorithm in our proposed decoding methods, we refer to it as the pseudo beam search. We use the term ``pseudo" because we restrict initial predictions to the $n$ most probable predictions, instead of spanning the entire vocabulary and for upcoming token predictions we restrict it to only top-1 prediction. For initial mask predictions, we constrain the search space to $(k \times m_s) \ll |V|$ which is much smaller than the entire vocabulary. For subsequent mask token predictions, we focus on the most probable outcomes. The source text $S$ undergoes pre-processing, as detailed in \Cref{eq:preprocess},

\begin{equation} \label{eq:preprocess}
  \begin{split}
  S  &= [t_1, t_2, ..., a_1, a_2, ..., a_r, ..., t_{n-1}, t_n]\\
  S' &= [t_1, t_2, ..., m_1, m_2, ..., m_r, ..., t_{n-1}, t_n]\\
  \end{split}
\end{equation}

where $m_i = m \quad \forall i=1,2,...,r$ and $m = \textit{<mask>}$ is the mask token. The generation/decoding strategy is an iterative process, and pseudo-formally, the generation of a candidate $c$ given the source $S$ and the answer $a$ is given in \Cref{eq:gen},

\begin{equation} \label{eq:gen} \small
    P(c_i|S,a) = 
        \begin{cases}
            \underset{\pi}{\mathrm{argmax}} \: P(m_1|S') & \text{if } i = 1 \\
             \underset{\pi}{\mathrm{argmax}} \: P(m_i|S'') & \text{otherwise} 
        \end{cases}
\end{equation}

where $\pi$ denotes the probability distribution over the vocabulary, and $S''$ is $S'$ where $m_j$ is replaced by $c_j$ at each step, i.e. $m_j := c_j \; \forall j<i$. Note that $c_{j<i}$ is the case for L2R decoding. 

In the generation phase, we propose two generation hyper-parameters; \emph{dispersion} and $n_{mask}$ (optional). The $n_{mask}$ parameter dictates the number of mask tokens that replace the removed answer tokens, in practice we used the number of answer tokens as $n_{mask}$ by default. The \emph{dispersion} parameter aims to enhance the diversity of generations by defining an interval for randomizing the number of mask tokens. In the pre-processing step (masking of the answer) we take \emph{dispersion} into consideration and we define an interval $[\max(n_{mask}-dispersion, 1), n_{mask}+dispersion]$ from which random numbers are drawn for multiple times. This results in contexts with varying numbers of replacement mask tokens, producing potentially multiple input variations.

By unmasking the mask tokens sequentially, we condition each subsequent token prediction on previously generated token(s) along with the context, due to $S''$ being updated at each step. While auto-regressive models generate predictions sequentially, MLMs are designed to predict each token simultaneously --- a key aspect of their pretraining objective. In our approach, we adapt an auto-regressive generation schema for masked Language Models (LM). This ensures that each generation is conditioned on previously generated \textbf{context}, leading to more natural and semantically coherent outputs. Upon multiple generations from multiple contexts (different mask token sizes), we rank these generations by their overall score which is the product of the probabilities along the generation, the score function is given in \Cref{eq:score}.

\begin{equation} \label{eq:score}
    T = \prod_{i=1}^r P(m_i|S'')
\end{equation}

Ranking all the candidates by \Cref{eq:score} is, however, not viable since the token size $r$ varies across generations (e.g. one context may use 3 mask tokens and another one may use 4). Applying the product of probabilities unfairly compares generations of different token lengths. To address this, we introduce a heuristic ranking score for candidates of any length using their probability scores: $T_{rank} = \sqrt[r]{T}$, which is a geometric mean of the probabilities of each token (in a step) along the generation. We opted for geometric mean as we have a multiplicative relation and it is more robust to outliers. In practice, we also performed experiments with harmonic mean, and harmonic mean, alternatively, can also be used in place of a geometric mean. The generation results by both averaging techniques are given in \Cref{tab:decodings}. The observations reveal slight differences between the averaging techniques. The final ranking of generated candidates is determined by $T_{rank}$ with higher values indicating better candidates.

\subsection{Generation Strategies}
\label{ssec:strat}

In the candidate generation step, we introduced the L2R and R2L decoding/generation strategies \cite{watanabe-sumita-2002-bidirectional}. Utilizing different decoding strategies enables the diversification of the candidate set outputs.

While these strategies are intuitive choices for generations, we recognize the potential for further generation strategies. Masked LMs consider the entire context during the generation phase, encompassing both the left and right surrounding contexts of the mask tokens. Building on this, we introduce the \textit{cocktail shaker decoding} (CTL) strategy, inspired by the cocktail shaker sort \cite{knuth1973art}). Notably, the CTL decoding strategy differs from the bidirectional decoding strategy proposed in \cite{watanabe-sumita-2002-bidirectional}. In particular, CTL alternates decoding the mask tokens with L2R and R2L step-by-step, rather than decoding L2R until the midway $\lceil \frac{m}{2} \rceil$ and R2L until the midway $\lfloor \frac{m}{2} \rfloor$ in reverse. For the detailed view, we refer readers to \Cref{apx:pipe}.

We also provided qualitative examples of generations from a SQuAD \cite{rajpurkar-etal-2016-squad} sample, showcasing different decoding strategies in \Cref{tab:decodings}. It can be easily seen that the differences between the decoding strategies are more significant compared to the averaging techniques.

\begin{table*}
\centering
\small
\begin{tabularx}{\textwidth}{c|X}
  \toprule
    Passage & Tesla was born on 10 July [O.S. 28 June] 1856 into a Serb family in the village of Smiljan, Austrian Empire (modern-day Croatia). His father, Milutin Tesla, was a Serbian Orthodox priest. Tesla's mother, Đuka Tesla (née Mandić), whose father was also an Orthodox priest,:10 had a talent for making home craft tools, mechanical appliances, and the ability to memorize Serbian epic poems. Đuka had never received a formal education. Nikola his eidetic memory and creative abilities to \textbf{his mother’s genetics} and influence. Tesla's progenitors were from western Serbia, near Montenegro.:12 \hfill\null \\
  \midrule
  Question & Who did Tesla credit for his abilities? \hfill\null \\\hline

  \makecell{Geometric Mean} & 
    \hfill \makecell[l]{\textbf{L2R}} 
    \hfill \makecell[l]{\textbf{R2L}} 
    \hfill \makecell[l]{\textbf{CTL}} \hfill\null \\\hline
  \makecell{1 \\ 2 \\ 3 \\ 4} & 
    \hfill \makecell[l]{the family's wealth \\ his mother's education \\ his own personal experience \\ his family tradition knowledge} 
    \hfill \makecell[l]{enhance his sense of power \\ have all his knowledge, power \\ preserve the Serbian tradition \\ be a combination of energy} 
    \hfill \makecell[l]{his own ageing, knowledge \\ his Serb religious upbringing \\ his own Serbian language skills \\ the influence of Serbian culture} \hfill\null \\\hline

  \makecell{Harmonic Mean} & 
    \hfill \makecell[l]{\textbf{L2R}} 
    \hfill \makecell[l]{\textbf{R2L}} 
    \hfill \makecell[l]{\textbf{CTL}} \hfill\null \\\hline
  \makecell{1 \\ 2 \\ 3 \\ 4} & 
    \hfill \makecell[l]{the family's wealth \\ his own personal experience \\ his family tradition knowledge \\ his mother's education} 
    \hfill \makecell[l]{enhance his sense of power \\ have all his knowledge, power \\ increase both his knowledge \\ be his sources of knowledge} 
    \hfill \makecell[l]{his Serb religious upbringing \\ his own ageing, knowledge \\  his own Serbian language skills \\ the influence of Serbian culture} \hfill\null \\

  \bottomrule
\end{tabularx}
\caption{Outputs from different decoding strategies and different averaging techniques. The answer is marked in bold font.}
\label{tab:decodings}
\end{table*}

\subsection{Distractor Selector (DS)}
\label{ssec:elimination}

After receiving the outputs from CSG, the pipeline proceeds with the equally crucial Distractor Selector (DS) phase. This phase aims to refine the candidate set by eliminating undesired candidates. While the CSG phase is adept at generating a diverse set of candidates, it doesn't guarantee the suitability of these candidates as true distractors. There are primarily two main reasons behind that:

\begin{itemize}
    \item The candidates might mirror the ground truth answer, either verbatim or in essence. This overlap will render a question invalid, as introduces multiple correct answers.
    \item The candidates themselves might mirror each other, such as presenting analogous distractors. While this doesn't outright invalidate a question, it undermines the quality of the distractors as a whole. By assuming only a single choice is the correct answer, a respondent can easily deduce duplicate choices as incorrect.
\end{itemize}

To address these challenges, our DS employs a language model fine-tuned on a downstream task of Natural Language Inference (NLI), sometimes also referred to as Recognizing Textual Entailment (RTE) \cite{dagan2005pascal}). This approach is markedly different from the DS in CDGP \cite{chiang-etal-2022-cdgp}, which relies on FastText \cite{bojanowski-etal-2017-enriching} word embedding model. For our experiments, we used a publicly available BART model\footnote{https://huggingface.co/geckos/bart-fined-tuned-on-entailment-classification} \cite{lewis-etal-2020-bart} fine-tuned on the NLI task. While this model is a three-way classifier for \emph{entailment}, \emph{neutral} and \emph{contradiction}, a two-way NLI model that distinguishes between \emph{entailment} and \emph{contradiction} could also be used.

The elimination process consists of two steps, directly addressing the two aforementioned challenges regarding the generated candidates. This phase is illustrated in \Cref{fig:disgem}, highlighted with a red background.

For the first step, the aim is to eliminate the candidates that share identical or similar meanings with the ground truth answer. To achieve this, we compare the source context $S$ (containing the ground truth answer) against the modified context where the ground truth answer is replaced with a candidate. Using the chosen NLI model, we then eliminate any candidate if the model's output is \textit{entailment}. This ensures that only candidates that either contradict or remain neutral in meaning compared to the ground truth answer are retained.

The second stage of the elimination process takes the filtered outputs of the first step as its starting point. In this phase, our goal is to ensure diversity among the candidates, eliminating any that share similar meanings, as this similarity would be undesirable within the final set of distractors. To tackle this challenge, we once again employ the NLI model. However, the difference this time is that we compare contexts populated with different candidates against each other. If the NLI model's result for any pair of contexts is \textit{entailment}, we discard the candidate with the lower score.

Following the second step of elimination, we are left with our finalized distractors. It is crucial to note that the NLI model we employed is trained on sentences rather than passages. Therefore, the input is formatted as $``<sentence_A><sentence_B>"$. To ensure thoroughness, we utilize a two-way entailment classification in our approach. That is, we provide the model with permutations of input: $``<sentence_A><sentence_B>"$ and $``<sentence_B><sentence_A>"$. In this setup, an \textit{entailment} is only when both results are \textit{entailment}.

\section{Experiments}
\label{sec:experiments}

We conducted two primary evaluations to assess our distractor generation method: Instruction-following Large Language Models (LLMs) evaluations and human evaluations. For LLM evaluations, we used SQuAD context/answer pairs to test different generation hyperparameters and compared the best to the previous work. For human evaluations, we used questions from the CLOTH dataset, comparing gold distractors, our generated distractors, and those from previous work, and measured human accuracy and quality ratings. Additional quantitative results from experiments on the CLOTH dataset are provided in Appendix \ref{apx:quant}.

\subsection{LLM Evaluations}

To evaluate the effectiveness of our distractor generation method, we used 100 random context/answer pairs from the SQuAD dataset. This dataset provides a diverse set of context passages and corresponding answers, which we utilized to create question/answer pairs. These pairs were then used to generate distractors with our method under different hyperparameter settings and compared against distractors generated by previous methods.

The evaluation involved the following steps:

\begin{enumerate}
    \item Question Selection: We randomly selected 100 pairs from the SQuAD dataset to ensure a representative sample. Using these pairs, we created fill-in-the-blank questions. This question set has been kept constant across different experiments.
    \item Distractor Generation: We generated distractors using our method with various hyperparameter settings ($n_{mask}$,  $dispersion$, decoding strategy) and compared them between each other and against distractors generated by previous methods.
    \item Best Set Selection by LLM: We employed ChatGPT-4o \cite{openai2024gpt4o} to select the best distractor set from the given sets (including ours and previous methods). Ties were allowed if the best options were close in quality.
\end{enumerate}

\subsubsection{Hyperparameter Experiments}
To identify the optimal settings for generating distractors, we experimented with different hyperparameters by systematically varying one parameter at a time while keeping the others constant. For distractor generation, we utilized RoBERTa\textsubscript{LARGE} \cite{liu2019roberta}, a widely recognized pre-trained transformer model.

\begin{itemize}
    \item First, we experimented with the number of mask tokens. We tested with $n_{\text{mask}} = 1$ and $n_{\text{mask}} = 0$ (where the number of mask tokens matched the number of answer tokens). For these experiments, the $dispersion$ parameter was set to $0$, and the decoding strategy was left-to-right (L2R).
    \item Next, we explored the impact of the $dispersion$ parameter. We experimented with $dispersion$ values of $0$, $1$, and $2$. In these experiments, we kept $n_{\text{mask}} = 0$ and continued using the L2R decoding strategy.
    \item Finally, we evaluated different decoding strategies to determine their effect on distractor quality. We compared left-to-right (L2R), right-to-left (R2L), and cocktail shaker (CTL) strategies. For these experiments, we set $n_{\text{mask}} = 0$ and $dispersion = 1$.
\end{itemize}

\begin{table}
\centering
\begin{tabular}{l|c}
\hline
\textbf{Parameter}                    & \textbf{Score} \\ \hline
\multicolumn{2}{c}{\textbf{Number of Mask Tokens}} \\ \hline
$n_{\text{mask}} = 0$                 & \textbf{57}             \\ 
$n_{\text{mask}} = 1$                 & 53             \\ \hline
\multicolumn{2}{c}{\textbf{Dispersion Parameter}} \\ \hline
$dispersion = 0$                        & 47             \\ 
$dispersion = 1$                        & \textbf{56}             \\ 
$dispersion = 2$                        & 50             \\ \hline
\multicolumn{2}{c}{\textbf{Decoding Strategy}} \\ \hline 
L2R (Left-to-Right)                   & 44             \\ 
R2L (Right-to-Left)                   & 29             \\
CTL (Cocktail Shaker)                 & \textbf{52}    \\ \hline
\end{tabular}
\caption{Scores for various hyperparameters used in distractor generation: the number of mask tokens ($n_{\text{mask}}$), $dispersion$ parameters, and decoding strategies. Each score reflects the preference of the referee LLM, ChatGPT-4o, for that hyperparameter in direct comparisons.}
\label{tab:hyperparameters}
\end{table}

\subsubsection{Results}

The results indicated several key findings, as summarized in \Cref{tab:hyperparameters}.

Experimenting with the number of mask tokens showed that using zero mask tokens, where the number of mask tokens matched the number of answer tokens, generally led to better distractor quality compared to using a single mask token.

When varying the dispersion parameter, we found that a moderate level of randomness in the number of mask tokens improved distractor quality. A dispersion parameter of 1 was more effective than having no dispersion or a higher level of dispersion.

Among the different decoding strategies, the Cocktail Shaker (CTL) strategy performed the best, followed by the Left-to-Right (L2R) and then the Right-to-Left (R2L) strategy. This indicates that the CTL strategy is more effective in generating high-quality distractors.

\begin{table}
\centering
\begin{tabular}{l|c}
\hline
\textbf{Method} & \textbf{Score} \\ \hline
CDGP            & 12             \\ 
DisGeM          & \textbf{88}             \\ \hline
\end{tabular}
\caption{Final scores for distractor generation methods: CDGP vs. DisGeM. Each score reflects the preference of the referee LLM, ChatGPT-4o, in direct comparisons.}
\label{tab:final_comparison}
\end{table}

Overall, the experiments demonstrated that the optimal configuration for generating high-quality distractors involves using a matching token count, a dispersion parameter of 1, and the Cocktail Shaker (CTL) decoding strategy. This configuration was compared to the previous work CDGP \cite{chiang-etal-2022-cdgp}, using their experiment setup from the CDGP codebase\footnote{https://github.com/AndyChiangSH/CDGP}. We conducted a final LLM comparison with their distractors versus our distractors on the  question set. The results, shown in \Cref{tab:final_comparison}, indicate that our method, DisGeM, significantly outperformed CDGP.

\subsection{Human Evaluation}
\label{subsubsec:results-human}

\begin{table}
\centering
\small
    \begin{tabular}{ p{2cm} | ccc }
        \toprule
        \textbf{Distractors} & \textbf{CDGP} & \textbf{DisGeM} & \textbf{GOLD} \\
        \midrule
        \textbf{Average} & 81.00 & 63.62 & 76.28 \\
        \textbf{Median} & 83.33 & 66.67 & 76.47\\
        \textbf{Min.} & 55.56 & 38.89	& 47.06 \\
        \textbf{Max.} & 94.44 & 83.33 & 100.00 \\
        \textbf{Std. Dev.} & 12.81 & 10.91 & 11.11 \\
    \bottomrule
    \end{tabular}
    \caption{Statistics for the correctness of the evaluators. ``GOLD" represents questions with ground truth distractors.}
    \label{tab:human-stat}
\end{table}

\begin{figure}
\centering
\includegraphics[width=\linewidth]{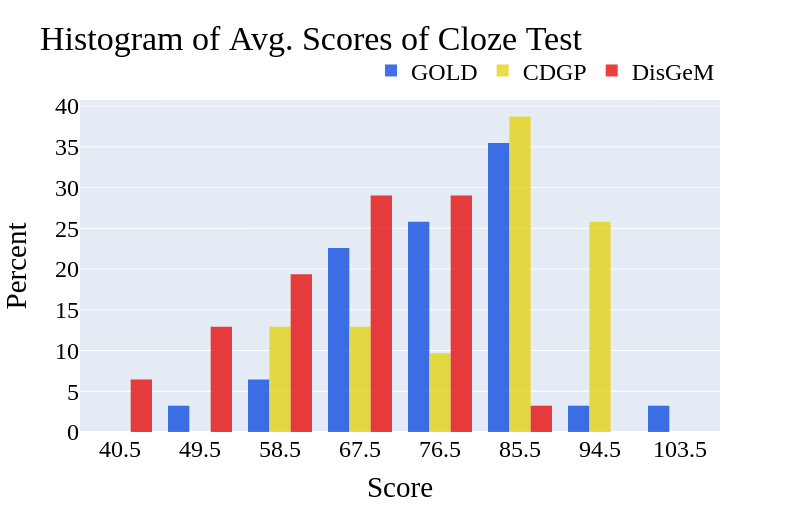}
\caption{Average human correctness of the cloze test grouped by the frameworks, ``GOLD" represents the questions with ground truth distractors.}
\label{fig:human_score}
\end{figure}

As previous works \cite{Ren_Zhu_2021, chiang-etal-2022-cdgp} have done, we also conducted human evaluation. For the evaluation process, we recruited 30 human evaluators, and we asked evaluators to take a cloze exam and rate the questions. Our evaluation process follows the evaluation setup of \citet{chiang-etal-2022-cdgp} with the following differences,

\begin{itemize}
    \item We prepared a cloze test with three passages randomly chosen from the CLOTH dataset, each passage is evenly split into three parts for ground truth, CDGP and DisGeM distractors.
    \item Unlike CDGP's ordering where the first 5 questions were ground truth and the last 5 questions were CDGP generations, we randomly shuffled the order of the questions to eliminate any possible bias from human evaluators.
    \item We included 17 questions in total for each distractor source (ground truth, CDGP, DisGeM) in three passages.
    \item We incorporated a simultaneous cloze test where we also gathered feedback regarding the quality and difficulty after each question. In this context, "difficulty" pertains to the extent of uncertainty or wavering among multiple answer choices (i.e. how well the distractors distract the evaluator). Here, we compounded the quality and difficulty assessment as a single feedback and asked the evaluators to rate on a Likert scale ranging from 1-5.  
\end{itemize}

\subsubsection{Results}

\begin{figure}
\centering
\includegraphics[width=\linewidth]{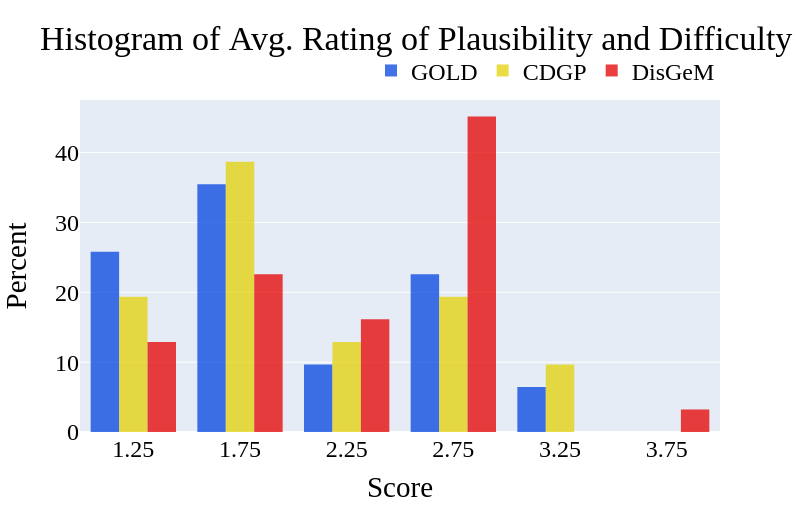}
\caption{Average rating of the quality and difficulty of the questions by the evaluators, ``GOLD" represents the questions with ground truth distractors.}
\label{fig:human_rating}
\end{figure}

We illustrate the histograms\footnote{We used the Plotly package \cite{plotly} for illustrations.} of the results of the cloze test and the results of the ratings by evaluators grouped by the three groups in \Cref{fig:human_score} and in \Cref{fig:human_rating} respectively. Detailed statistics are given in \Cref{tab:human-stat}. The histograms reveal a noticeable trend in the rate of correctness of evaluators is comparatively lower on DisGeM's questions than on those from GOLD and CDGP. Also, more evaluators voted higher for DisGeM in quality and difficulty rating of questions/distractors compared to GOLD and CDGP. 

For the assessment of the statistical significance we also conduct a Student's t-test to show that the question group of DisGeM is slightly harder than that of GOLD (i.e. the ground truth distractors). \citet{chiang-etal-2022-cdgp} has already stated in their human evaluation that their questions/distractors are slightly easier from the GOLD. Our human evaluation results are also parallel to this conclusion. Hence, we moved forward to compare DisGeM and GOLD groups. We conducted Student's t-test \cite{student1908probable} under the null hypothesis $\mu_{disgem} - \mu_{gold} < 0$, and get the p-value 0.999, and thus we have no significant evidence to reject the null hypothesis. Also, the extremity of the p-value suggests that the null hypothesis is likely to be the case.

\section{Discussion}
\label{sec:discussion}

The evaluation results from both LLM and human evaluations provide significant insights into the effectiveness of our distractor generation method, DisGeM, compared to previous methods. In the LLM evaluations, we observed that the optimal configuration of a matching token count, a dispersion parameter of 1, and the Cocktail Shaker (CTL) decoding strategy showed a significant improvement over CDGP, as indicated by the final comparison scores.

In human evaluations, the results were similarly encouraging. The cloze test scores revealed that evaluators had a slightly lower correctness rate on DisGeM's questions compared to GOLD and CDGP. This lower correctness rate indicates that DisGeM's distractors were more effective at introducing uncertainty and challenging the evaluators. The quality and difficulty ratings from evaluators were also higher for DisGeM, suggesting that our method produces distractors that are perceived as more engaging and effective at creating uncertainty.

\section{Conclusion}
\label{sec:conclusion}

We propose a training-free dual-stage distractor generation framework, DisGeM. Our framework possesses two main features extending previous works. DisGeM in particular,

\begin{itemize}
    \item is a training-free framework and does not necessitate fine-tuning on a particular dataset though one can opt for various reasons (e.g. domain adaptation).
    \item has a novel multi-word distractor generation algorithm.
    \item has far less amount of generation hyper-parameters that need to be tuned by users/researchers.
    \item incorporate strategies (e.g. decoding, multi-token generation) to achieve natural variation in the generations.
\end{itemize}

Also, due to the training-free nature of our framework can directly be used on any passage without the effort of additional training/fine-tuning, and hence saves a lot of time. Additionally, it alleviates the heavy hyper-parameter tuning stage. 

Our evaluations with LLM and human participants demonstrate DisGeM's effectiveness. LLM evaluations showed DisGeM outperforms previous methods. Human evaluations confirmed this with lower correctness rates and higher quality and difficulty ratings, indicating more effective and engaging distractors.

\section*{Limitations}
\label{sec:limitations}

We have so far laid out results along with a discussion on them. Although the results were promising there are several limitations to the proposal. Like its predecessors \cite{Ren_Zhu_2021, chiang-etal-2022-cdgp}, DisGeM is also proposed as a distractor generation framework for extractive MCQs and for English only, and experiments are conducted according to the extractive MCQ setup. The framework on its own is not capable of working on an abstractive MCQ setting, where the answer is not necessarily a span/substring in the context, as it depends on the masking of the answer span in the context. Nonetheless, this might be achieved in future research that enhances the CSG phase.

Another limitation of DisGeM similar to the prior studies \cite{Ren_Zhu_2021, chiang-etal-2022-cdgp} is that there is no control of the difficulty level for generated distractors. Yet, this might be a good research direction for related future studies as it may have a positive impact on real-life quiz applications.


 
Finally, we would like to point out that we did not conduct any comparison with the instruction following LLMs or PLMs having parameters many times more than our PLMs in experiments (e.g. \cite{brown2020language, ouyang2022training, touvron2023llama, openai2024gpt4o}). The primary reason for not comparing with those kinds of models was the absence of a standardized methodology for employing such models as baseline performance. The performance of these models can depend on prompt engineering, model versions, and whether the approach is zero-shot or fine-tuned. Also, general models like those may generate `hallucinated' content, which requires additional layers of validation. Furthermore, a kind of comprehensive evaluation is needed to compare against those models which requires significant time investment. 

Apart from the limitations discussed, there are also potential risks that could primarily affect education. While the proposed framework may be promising in experiments, these experiments can not provide a full-picture overview. Using this framework in real-life quizzes and tests may not give a correct assessment of the test takers. Thus, the framework should be used under the guidance of a human expert (e.g. teacher) in those scenarios. This is further amplified by our lack of control over how difficult the distractors are.


\bibliography{custom}

\appendix

\section{Experimental Setup}
\label{apx:experiment}

The hardware specification that is used for conducting experiments are given as follows:

\begin{itemize}
\small
    \item CPU: AMD Ryzen 9 7900X 12-Core Processor
    \item GPU: NVIDIA RTX 3090 Ti
\end{itemize}

\section{Pipeline}
\label{apx:pipe}

We present the detailed view and the algorithm on our distractor generation pipeline. We describe pseudo-code algorithms for both CSG and DS phases.

\subsection{Candidate Set Generator (CSG)}
Our CSG phase of the framework is detailed in \Cref{alg:csg}. This phase highlights how our CSG differs from the previous work, CDGP. The candidate set generation phase utilizes methods such as randomly sampling the number of tokens with provided hyper-parameters (e.g. $dispersion$, $n_{mask}$). Our pseudo-beam search operates with a fixed window size of 1, after the first step of $k \times m_s$ generations. In the algorithm, we designed the search multiplier $m_s$ to allow users to adjust it based on their preferred trade-off between speed and fidelity. In practice, we set $m_s$ to 10 for the single-mask cases (like CLOTH outputs) when there is only one token and set it to 7 other scenarios.

\begin{algorithm*}
\small
\SetAlgoLined
\caption{Overall CSG pipeline. Indexing is assumed to be starting from 0.}
\label{alg:csg}
\KwData{$P$ passage, $A$ answer, $M_{PLM}$ Language Model (pretrained with MLM task), $T_{PLM}$ tokenizer of $M_{PLM}$, $k$ number of distractors, $s$ decoding strategy, $d$ dispersion, $m_s$ search multiplier, $n_{mask}$ number of mask tokens, $avg$ averaging technique (geometric or harmonic)}
\KwResult{$CS$ candidate set}
$P_T \gets T_{PLM}(P)$ \tcp*{tokenize passage}
\If{$n_{mask} == 0$}{
    $n_{mask} \gets |A_T|$
}
$N \gets  draw\_three(a=\max(n_{mask}-dispersion, 1), b=n_{mask}+dispersion, replace=False)$\;
$CS \gets [ \; ]$\;
$RS \gets [ \; ]$\;
\For{$i \gets 0$ to $|N|-1$}{
    $M_T \gets [<mask>_0,\dots, <mask>_{N_i - 1}]$\;
    $P_T \gets replace(A_T, M_T, P_T)$ \tcp*{replace answer tokens by <mask> tokens}
    $m \gets k * m_s$\;
    $C \gets empty\_array(m, |M_T|)$\;
    $R \gets empty\_array(m, |M_T|)$\;
    $P'_T \gets tile\_array(P_T, m)$ \tcp*{$P'_T$ is a 2 dim array of size $(m, |P_T|)$}
    \For{$j \gets 0$ to $|M_T|-1$}{ 
        \uIf{$j==1$}{
            $t = m$\;   
        }\uElse{
            $t = 1$\;
        }
        $C_{\boldsymbol{\cdot} j}, R_{\boldsymbol{\cdot} j} \gets M_{PLM}(P'_T, token=M_{T_j}, top_k=t)$ \tcp*{get $t$ predictions for $j^{th}$ token of $M_T$}  
        $P'_T \gets replace(M_{T_j}, C_{\boldsymbol{\cdot}j}, P'_T)$\;

    }
    $C \gets concat\_strings(C, axis=1)$\;
    $R_{prod} \gets product(R, axis=0)$\;
    $C \gets sort(C, by=R_{prod}, descending=True)$\;
    $add(C, CS)$\;
    $add(R, RS)$ \tcp*{RS and CS are 2 dim arrays with shapes $(3m,|M_T|)$}
}
$R_{avg} \gets average(RS, type=avg, axis=0)$\;
$CS \gets sort(C, by=R_{avg}, descending=True)$\;
\Return{$CS$}\;
\end{algorithm*}

\subsection{Distractor Selector (DS)}
The pipeline details for the distractor selection phase can be found in  \Cref{alg:ds}. As previously mentioned, our approach utilizes a two-way entailment check. In our context, checking for entailment from both the answer to the distractor and the distractor to the answer is undesirable for distractor generation.

\begin{algorithm*}
\small
\SetAlgoLined
\caption{Overall DS pipeline. Indexing is assumed to start from 0. $|\cdot|$ refers to the norm/length of a component.}
\label{alg:ds}
\KwData{$S$ sentence, $A$ answer, $C$ candidates by CSG, $M_{NLI}$ NLI model, $k$ number of distractors}
\KwResult{$D$ distractors}
\For{$n \gets |C|-1$ to $0$}{
    \tcp{remove candidates entailing with the answer}
    $P_d \gets replace(C_n, A, P)$\;
    $check \gets (M_{NLI}(P_d,P), M_{NLI}(P_d,P))$ \tcp*{two-way entailment check}
    \If{$check == (entailment, entailment)$}{
        $remove\_item(C, index=n)$
    }
}
$increment \gets True$\;
$i_{kept} \gets 1$\;
\While{$i_{kept} < k < |C|$}{
    \tcp{remove candidates entailing within (among selected distractors)}
    \For{$j \gets 1$ to $i_{kept}$}{
        $P_{d1} \gets replace(C_j, A, P)$\;
        $P_{d2} \gets replace(C_{i_{kept}}, A, P)$\;
        $check \gets (M_{NLI}(P_{d1},P_{d2}), M_{NLI}(P_{d2},P_{d1}))$ \tcp*{two-way entailment check}
        \If{$check == (entailment, entailment)$}{
            $increment \gets False$\;
            $remove\_item(C, index=i_{kept})$\;
            \Break
        }
        $increment \gets True$\;
    }
    \If{$increment$}{
        $i_{kept}++$\;
    }
}
\Return{$get\_top\_k(C,k)$}\;
\end{algorithm*}

\subsection{Decoding Strategies}
\begin{figure*}
\begin{subfigure}[t]{\linewidth}
\centering
\includegraphics[width=\linewidth]{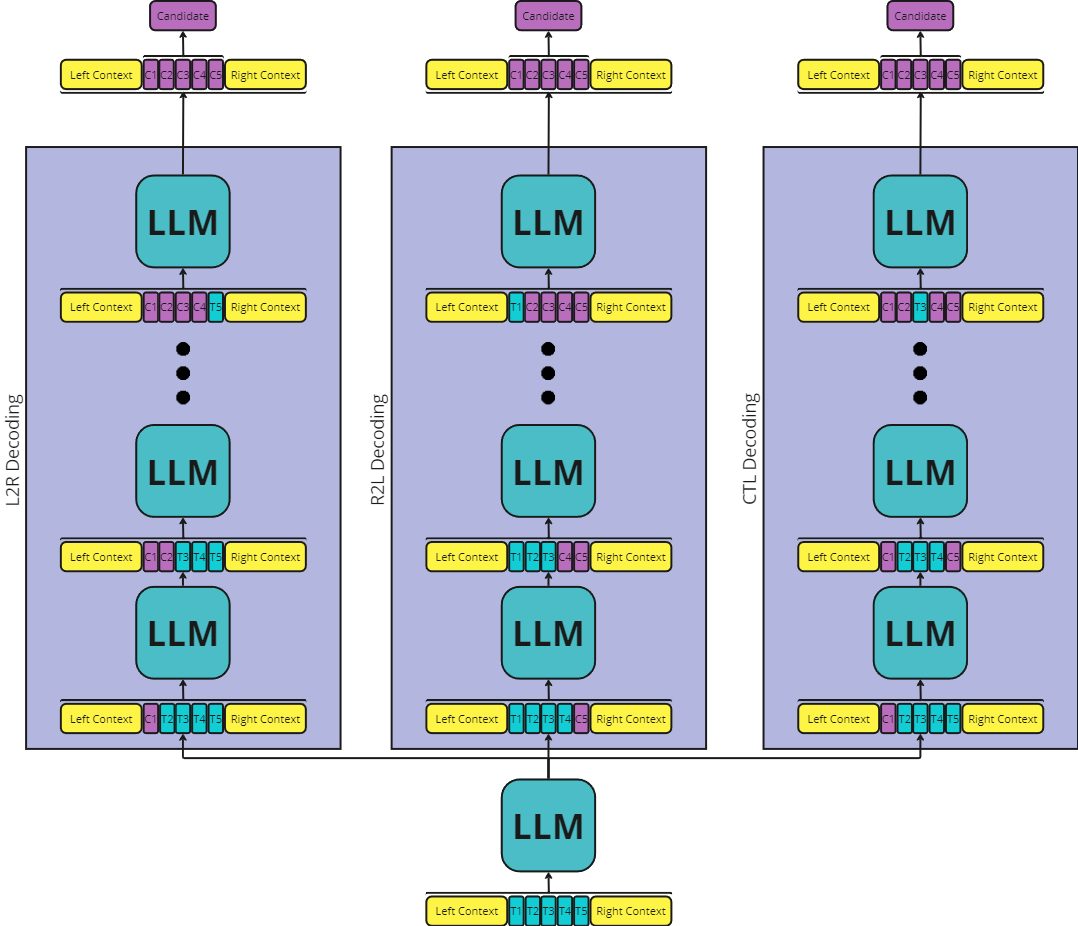}

\end{subfigure}\hfill
\caption{Different strategies proposed for generating candidates. The prediction orders for mask tokens are \textbf{(left) L2R} 1-2-3-4-5, \textbf{(middle) R2L} 5-4-3-2-1, \textbf{(right) CTL} 1-5-2-4-3. ``T" in blue refers to mask tokens. ``C" in purple refers to the candidate tokens.}
\label{fig:strats}
\end{figure*}

We have conducted experiments and discussed the outputs in \Cref{sec:method}. The schema in \Cref{fig:strats} illustrates various generation strategies, including our newly proposed cocktail shaker (CTL) decoding strategy.

\section{Quantitative Results}
\label{apx:quant}
Following the previous works \cite{Ren_Zhu_2021, chiang-etal-2022-cdgp} we evaluated our framework with automated metrics. We used the experiment setup from CDGP \cite{chiang-etal-2022-cdgp} codebase\footnote{https://github.com/AndyChiangSH/CDGP}.

\subsection{Dataset}
\label{subsec:dataset}

Following \cite{chiang-etal-2022-cdgp} we evaluated our framework on CLOTH dataset \cite{xie-etal-2018-large}. We used the instances from the test split (high) of the CLOTH dataset to evaluate our framework. Since we did not conduct any fine-tuning, we only focused on the test split.

CLOTH is a dataset curated by teachers and consists of passages/paragraphs containing cloze-style questions where the answers are one word and with 4 options in MCQ style. The distractors/options in MCQ The statistics regarding the test split (high) of CLOTH are given in \Cref{tab:cloth-stat}.

\subsection{Evaluation Metric}
\label{subsec:exp-eval-metric}

We adhere to the evaluation metric setup established in \cite{Ren_Zhu_2021, chiang-etal-2022-cdgp}, assessing Precision (P@1), F1 score (F1@3), Mean Reciprocal Rank (MRR@10), and Normalized Discounted Cumulative Gain (NDCG@10).

\subsection{Experimental Setup}
\label{subsec:exp-setup}

We used BERT\textsubscript{LARGE} \cite{devlin-etal-2019-bert} and RoBERTa\textsubscript{LARGE} \cite{liu2019roberta} as PLMs for the CSG phase to conduct our experiments following \cite{chiang-etal-2022-cdgp}. Since CLOTH dataset answers and distractors consist of a single word, we set $dispersion=0$, $n_{mask}=1$, $k=10$ and $m_s = 7$. Note that forcing $n_{mask}=1$ does not truly demonstrate the capabilities of our framework as our framework is capable of generating multi-word distractors. We set L2R decoding and used the geometric mean for averaging. Hardware specifications are detailed in \Cref{apx:experiment}.

\begin{table}
\centering
\small
    \begin{tabular}{ l | p{2cm} }
        \toprule
        \textbf{Dataset} &  \textbf{CLOTH (High/Test)} \\
        \midrule
        \textbf{\# passages} & 478 \\
        \textbf{Avg. \# question per passage} & 17.41 \\
        \textbf{Avg. \# sentence} & 18.92 \\
        \textbf{Avg. \# words} & 365.1\\
    \bottomrule
    \end{tabular}
    \caption{Statistics of the CLOTH dataset test split (high).}
    \label{tab:cloth-stat}
\end{table}

\subsection{Results}
\label{subsec:results}
In this section, we lay out the results either quantitative or qualitative that are obtained from the experiments with aforementioned datasets.

\begin{table*}
\small
\centering
    \begin{tabular}{ p{3.5cm} | p{1cm} p{1cm} p{1.5cm} p{1.6cm} }
        \toprule
        \textbf{Models} & \textbf{P@1} & \textbf{F1@3} & \textbf{MRR@10} & \textbf{NDCG@10} \\
        \midrule
        CDGP (BERT) & \textbf{18.50} & \textbf{13.80} &  \textbf{29.96} & \textbf{37.82} \\
        CDGP (RoBERTa) & 10.50 & 9.83 & 20.42 & 28.17 \\
        \midrule
        DisGeM (BERT) & 14.00 & 7.67 & 19.03 & 22.94 \\
        DisGeM (RoBERTa) & 8.00 & 6.00 & 14.80 & 19.92 \\
    \bottomrule
    \end{tabular}
    \caption{Automated metrics from the evaluation on CLOTH dataset test split (high) and comparison with CDGP \cite{chiang-etal-2022-cdgp}.}
    \label{tab:cloth}
\end{table*}

\begin{table*}
\small
\centering
    \begin{tabular}{ p{3cm} | p{1cm} p{1cm} p{1.5cm} p{1.6cm} }
        \toprule
        \textbf{Models} & \textbf{P@1} & \textbf{F1@3} & \textbf{MRR@10} & \textbf{NDCG@10} \\
        \midrule
        BERT (P/FM) & \textbf{14.00} & \textbf{7.67} & \textbf{19.03} & \textbf{22.94} \\
        BERT (P/G) & 12.00 & 6.67 & 16.90 & 20.89 \\
        BERT (S/G) & 11.00 & 5.33 & 15.01 & 18.60 \\
    \bottomrule
    \end{tabular}
    \caption{Automated metrics from the evaluation on CLOTH dataset test split (high) utilizing different techniques on DisGeM. Model input is \textbf{(P)} the whole passage \textbf{(S)} the sentence containing the blank/question. If the input context has more blanks other than the question at hand \textbf{(FM)} they are pre-filled using a PLM, \textbf{(G)} they are pre-filled with gold answers.}
    \label{tab:cloth-disgem}
\end{table*}

The performance of the framework and comparison with CDGP on the CLOTH dataset is reported in \Cref{tab:cloth}. We followed the same experiment setup as CDGP. It can be seen that despite not using a fine-tuned PLM on CSG, DisGeM's results are compatible with CDGP's. Moreover, the trade-off for performance vs. fine-tuning may be worth using DisGeM (e.g. for the RoBERTa case).

We also conduct a study with different input settings whose results are reported in \Cref{tab:cloth-disgem}. We tried supplying the whole passage to the model and also supplying only the sentence with the question at hand. Probably due to the pre-training nature of BERT \cite{devlin-etal-2019-bert}, the generations with passages are closer to the ground truth distractors. Furthermore, we also tried pre-filling the blanks other than the question at hand with a PLM compared to pre-filling with the ground truth distractors. Interestingly enough, when the blanks are pre-filled with PLMs (we used RoBERTa-large \cite{liu2019roberta}) the generations are closer to the ground truth. We did not conduct a pre-filling comparison for sentence inputs as the majority of the sentences have a single blank, and used ground truth answers only for those that have more than 1 blank.

\subsection{Discussion}

In the metric evaluations, the reported automated metrics either cannot reach SOTA or are comparable to some end for some variants (e.g. CDGP RoBERTa results in \Cref{tab:cloth}). These results are not surprising for several reasons. First of all, our framework is training-free and we do not fine-tune PLMs for specific datasets (e.g. CLOTH) whereas in CDGP's framework, the results are obtained by PLMs that are fine-tuned on the datasets. Hence, it's expected that CDGP's framework is better at matching the distractors of that particular domain. We do not want to state that CDGP's or DisGeM's better in comparison by only interpreting the automatic metrics on CLOTH datasets since this kind of evaluation measures that the system is capable of matching the gold distractors written by people for that dataset. As \citet{chiang-etal-2022-cdgp} also states, the current evaluation on the test dataset may not truly represent the quality of the distractor generation system as a mismatch with the ground truth distractors does not necessarily indicate the infeasibility of the generated distractor.

It's also worth mentioning that the CLOTH dataset consists of single-word answers and distractors, which somewhat restricts our system from demonstrating its capabilities to the fullest extent. 

\section{Qualitative Analysis}
\label{apx:qual}

In this section, we present qualitative samples from our framework's outputs, accompanied by commentary. 

For the qualitative analysis, we have selected examples from the SQuAD dataset. It is important to note that the SQuAD dataset is primarily designed as an extractive question-answering dataset, where the answers are directly extracted as spans from the context. While primarily for question answering, it can be used for tasks other, such as question generation. Additionally, we'd like to emphasize that our framework, \textit{DisGeM}, focuses solely on the provided context and not any specific question.

Our distractor generation results, obtained using parameters $dispersion=1$, L2R decoding, and $n_{mask}=0$ (i.e. equals the length of the answer tokens), are showcased in \Cref{tab:superbowl}, \Cref{tab:tesla1} and \Cref{tab:tesla2} with comparison to CDGP results. To give a realistic representation of MCQs, we randomly positioned the correct answer among the choices without any particular order. Note that CDGP is primarily designed for the use case where the PLMs are first fine-tuned, and these fine-tuned models are then utilized to generate candidates. However, it is important to note that such fine-tuning is specific to multiple-choice question-answering datasets because the training objective relies on ground truth distractors. Contrastingly, the SQuAD dataset consists solely of contexts, questions and answers. To ensure a balanced comparison, we included results from two CDGP outputs: the BERT model fine-tuned on CLOTH (referred to as CDGP (F)) and the pre-trained BERT base (referred to as CDGP (P)), which we also use for DisGeM. 

Our qualitative analyses affirm that the candidates generated using the entire passage (as with DisGeM) are significantly more effective than those produced when only the sentence is provided (as in the case of CDGP). In our analyses, we refer to this as ``context awareness". Specifically, we use ``shallower context awareness" (SCA) when the model input is limited to a sentence and ``wider context awareness" (WCA) when it encompasses an entire passage. 

\subsection{Passage 1 - Super Bowl 50}

\begin{table*}
\small
\centering
\begin{tabularx}{\linewidth}{c|X}
  \toprule
    Passage-1 &   Super Bowl 50 was an American football game to determine the champion of the National Football League (NFL) for the \qblank{1} season. The American Football Conference (AFC) champion \qblank{2} defeated the National Football Conference (NFC) champion \qblank{3} 24–10 to earn their third Super Bowl title. The game was played on February 7, 2016, at Levi's Stadium in the San Francisco Bay Area at Santa Clara, California. As this was the 50th Super Bowl, the league emphasized the "golden anniversary" with various gold-themed initiatives, as well as temporarily suspending the tradition of naming each Super Bowl game with Roman numerals (under which the game would have been known as "Super Bowl L"), so that the logo could prominently feature the Arabic numerals 50. \hfill\null \\
  \midrule
  Question-1.1 & Super Bowl 50 decided the NFL champion for what season? \hfill\null \\\hline

  \makecell{Choices} & 
    \hfill \makecell{DisGeM} 
    \hfill \makecell{CDGP (F)}
    \hfill \makecell{CDGP (P)}\hfill\null \\\hline
  \makecell{} & 
    \hfill \makecell{A. 2016 \\ B. 2014 \\ C. 2017 \\ \textbf{D. 2015}} 
    \hfill \makecell{A. 2010 \\ B. 2014 \\ C. 2017 \\ \textbf{D. 2015}}
    \hfill \makecell{A. 1950 \\ B. 1949 \\ C. 1951 \\ \textbf{D. 2015}}
    \hfill\null \\\hline
  Question-1.2 & Which NFL team represented the AFC at Super Bowl 50? \hfill\null \\\hline
  \makecell{Choices} & 
    \hfill \makecell{DisGeM} 
    \hfill \makecell{CDGP (F)}
    \hfill \makecell{CDGP (P)} \hfill\null \\\hline
  \makecell{} & 
    \hfill \makecell{A. Patriots \\ \textbf{B. Denver Broncos} \\ C. Miami Dolphins \\ D. Houston Texans} 
    \hfill \makecell{A. denver \\ \textbf{B. Denver Broncos} \\ C. eventually \\ D. .} 
    \hfill \makecell{A. Patriots \\ \textbf{B. Denver Broncos} \\ C. Steelers \\ D. Colts}
    \hfill\null \\\hline
  Question-1.3 & Which NFL team represented the NFC at Super Bowl 50? \hfill\null \\\hline
    \makecell{Choices} & 
    \hfill \makecell{DisGeM} 
    \hfill \makecell{CDGP (F)}
    \hfill \makecell{CDGP (P)} \hfill\null \\\hline
  \makecell{} & 
    \hfill \makecell{\textbf{A. Carolina Panthers} \\  B. 49ers \\ C. Philedelphia Eagles \\ D.  Miami Dolphins} 
    \hfill \makecell{\textbf{A. Carolina Panthers} \\ B. carolina \\ C. , \\ D. .} 
    \hfill \makecell{\textbf{A. Carolina Panthers} \\ B. Denver \\ C. Broncos \\ D. Colts}
    \hfill\null\\
  \bottomrule
\end{tabularx}
\caption{Several SQuAD examples with context, question, answer and generated distractors. For CDGP outputs are with \textbf{(F)} BERT model fine-tuned on the CLOTH dataset \textbf{(P)} pre-trained BERT model.}
    \label{tab:superbowl}
\end{table*}

On \Cref{tab:superbowl} we selected the passage from the ``Super\_Bowl\_50" article\footnote{https://rajpurkar.github.io/SQuAD-explorer/explore/1.1/dev/Super\_Bowl\_50.html}. The passage is about Super Bowl 50 in particular, mentioning related information such as the competing teams, and when and where the game is played. 

The generated distractors on Question-1.1, DisGeM successfully generated distractors that are close to the gold answer, ``2015". Given that the context mentions Super Bowl 50 was played in 2016, ``2016" emerges as a plausible distractor. Options 2014 and 2017 are also closer years to 2015. While the outputs from CDGP (F) closely resemble those of DisGeM, option (A) deviates slightly from other choices. Conversely, the quality of outputs from CDGP (P) is noticeably lower compared to both DisGeM and CDGP (F). This case stands as an effective illustration of the beneficial impact of WCA over SCA.

Upon examining the distractors for Question 1.2, it becomes evident that the distractors generated by CDGP (F) are less effective than the rest. It generated ``denver" as a distractor, which is functionally identical to the correct answer, thus invalidating the question. Both DisGeM and CDGP (P) are comparable, though DisGeM's distractors appear slightly more plausible, given its multi-word generation capability. Other than that both DisGeM and CDGP (P) successfully generated distractors of AFC teams.

For Question-1.3, DisGeM's outputs outshine those from both CDGP (F) and CDGP (P). This time, unlike Question-1.2, one of the DisGeM's distractors, ``Miami Dolphins", is an AFC team, but not NFC. Both CDGP (F) and CDGP (P) generated undesirable distractors. For CDGP (F), there is only a single acceptable string as a candidate, but once again it is extremely similar to the correct answer. On the other hand, CDGP (P) generated ``Denver" and ``Broncos" as distractors which refer to the same team ``Denver Broncos" making the distractors suboptimal.

\subsection{Passage 2 - Nikola Tesla}

\begin{table*}
\small
\centering
\begin{tabularx}{\linewidth}{c|X}
  \toprule
    Passage-2 & Tesla was renowned for his achievements and showmanship, eventually earning him a reputation in popular culture as an archetypal \qblank{1}. His patents earned him a considerable amount of money, much of which was used to finance his own projects with varying degrees of success. He lived most of his life in a series of \qblank{2}, through his retirement. Tesla died on 7 January 1943. His work fell into relative obscurity after his death, but in 1960 the General Conference on Weights and Measures named the \qblank{3} the tesla in his honor. There has been a resurgence in popular interest in Tesla since the 1990s.\\\hline
    Question-2.1 & What was Tesla's reputation in popular culture? \hfill\null \\\hline
    \makecell{Choices} & 
    \hfill \makecell{DisGeM} 
    \hfill \makecell{CDGP (F)}
    \hfill \makecell{CDGP (P)} \hfill\null \\\hline
    \makecell{} & 
    \hfill \makecell{A. inventor \\  B. genius \\ \textbf{C. mad scientist} \\ D.  electric genius} 
    \hfill \makecell{A. actor \\ B. artist \\ \textbf{C. mad scientist} \\ D. writer} 
    \hfill \makecell{A. figure \\ B. hero \\ \textbf{C. mad scientist} \\ D. inventor} 
    \hfill\null \\\hline
    Question-2.2 & Where did Tesla live for much of his life? \hfill\null \\\hline
    \makecell{Choices} & 
    \hfill \makecell{DisGeM} 
    \hfill \makecell{CDGP (F)}
    \hfill \makecell{CDGP (P)} \hfill\null \\\hline
    \makecell{} & 
    \hfill \makecell{A. small retirement homes \\  B. homes worldwide \\ \textbf{C. New York hotels} \\ D.  hospitals} 
    \hfill \makecell{A. hospitals \\ B. parks \\ \textbf{C. New York hotels} \\ D. hotels} 
    \hfill \makecell{A. houses \\ B. homes \\ \textbf{C. New York hotels} \\ D. apartments} 
    \hfill\null \\\hline
    Question-2.3 & What was named “The Tesla” in his honor? \hfill\null \\\hline
    \makecell{Choices} & 
    \hfill \makecell{DisGeM} 
    \hfill \makecell{CDGP (F)}
    \hfill \makecell{CDGP (P)} \hfill\null \\\hline
    \makecell{} & 
    \hfill \makecell{A. new precision \\ \> measurement device \\  B. new ``universal motor" \\ \textbf{C. SI unit of} \\ \> \textbf{magnetic flux density} \\ D.  first ``universal" electric car} 
    \hfill \makecell{A. model \\ B. unit \\ \textbf{C. SI unit of} \\ \> \textbf{magnetic flux density} \\ D. field}
    \hfill \makecell{A. device \\ B. instrument \\ \textbf{C. SI unit of} \\ \> \textbf{magnetic flux density} \\ D. asteroid}
    \hfill\null \\
  \bottomrule
\end{tabularx}
\caption{Several SQuAD examples with context, question, answer and generated distractors.For CDGP outputs are with \textbf{(F)} BERT model fine-tuned on the CLOTH dataset \textbf{(P)} pre-trained BERT model.}
    \label{tab:tesla1}
\end{table*}

On \Cref{tab:tesla1} we selected the passage from the ``Nikola\_Tesla" article\footnote{https://rajpurkar.github.io/SQuAD-explorer/explore/1.1/dev/Nikola\_Tesla.html}. The passage, Passage 2, is a paragraph about the work life of Nikola Tesla and the events that took place posthumously. 

Analyzing the outcomes for Question-2.1, DisGeM's distractors for the answer ``mad scientist" aptly capture Tesla's reputation in popular culture. Conversely, CDGP (F) and CDGP (P) offered less fitting alternatives, with ``mad scientist" being the most appropriate choice among their options though DisGeM and CDGP (P) have a common plausible distractor, ``inventor". Nonetheless, DisGeM's ``genius" and ``electric genius" distractors are close to each other in meaning which may be undesirable. This underlines DisGeM's ability to generate context-aware distractors. 

Moving to Question 2.2, the comparison highlights both CDGP (F) and CDGP (P) as producing less plausible distractors. Remarkably, CDGP (F) generated ``hotels" as a distractor, which directly agrees with the answer ``New York hotels". Besides, CDGP (P) distractors are all very similar in meaning, significantly reducing the quality of the overall question and choices. In contrast, DisGeM yielded options that are more contextually feasible, especially option (A), ``small retirement homes", which is quite relatable with the context of retirement.

In the case of Question 2.3, the results show that DisGeM's distractors capture the nuances of the given answer, contrasting with the more generalized options from CDGP (F) and CDGP (P). This allows the distractors to compete more effectively with the answer ``SI unit of magnetic flux density". Both CDGP (F) and CDGP (P) distractors substantially differ from the correct answer. Placing a lengthy, detailed answer next to short and general distractors significantly compromises the integrity of the question. Conversely, DisGeM showcases its proficiency by creating distractors that resonate with the answer well, although Option (D), ``first ``universal" electric car", introduces an element of whimsy that feels slightly out of place.

\subsection{Passage 3 - Nikola Tesla}

\begin{table*}
\small
\centering
\begin{tabularx}{\linewidth}{c|X}
  \toprule
  Part & 
    \hfill SQuAD Samples \hfill\null \\
  \midrule
    Passage-3 & Tesla was born on 10 July [O.S. 28 June] 1856 into a Serb family in the village of Smiljan, Austrian Empire (modern-day \qblank{1}). His father, Milutin Tesla, was a Serbian Orthodox priest. Tesla's mother, Đuka Tesla (née Mandić), whose father was also an Orthodox priest,:10 had a talent for making home craft tools, mechanical appliances, and the ability to memorize Serbian epic poems. Đuka had never received a formal education. Nikola credited his eidetic memory and creative abilities to \qblank{2} and influence. Tesla's progenitors were from western Serbia, near Montenegro.:12\\\hline
    Question-3.1 & What modern-day country was Tesla born in? \hfill\null \\\hline
    \makecell{Choices} & 
    \hfill \makecell{DisGeM} 
    \hfill \makecell{CDGP (F)}
    \hfill \makecell{CDGP (P)}\hfill\null \\\hline
    \makecell{} & 
    \hfill \makecell{A. Serbia \\  B. Montenegro \\ \textbf{C. Croatia} \\ D.  Bosnia \\ \> Herzegovina} 
    \hfill \makecell{A. france \\ B. china \\ \textbf{C. Croatia} \\ D. hungary} 
    \hfill \makecell{A. Croatia \\ B. Serbia \\ \textbf{C. Croatia} \\ D. Slovenia}
    \hfill\null \\\hline
    Question-3.2 & Who did Tesla credit for his abilities? \hfill\null \\\hline
    \makecell{Choices} & 
    \hfill \makecell{DisGeM} 
    \hfill \makecell{CDGP (F)}
    \hfill \makecell{CDGP (P)} \hfill\null \\\hline
    \makecell{} & 
    \hfill \makecell{A. the family’s wealth \\  B. is own personal experience \\ \textbf{C. his mother’s genetics} \\ D. his mother’s education} 
    \hfill \makecell{A. study \\ B. change \\ \textbf{C. his mother’s genetics} \\ D. adapt}
    \hfill \makecell{A. him \\ B. inspiration \\ \textbf{C. his mother’s genetics} \\ D. success} 
    \hfill\null \\\hline
  \bottomrule
\end{tabularx}
\caption{Several SQuAD examples with context, question, answer and generated distractors. For CDGP outputs are with \textbf{(F)} BERT model fine-tuned on the CLOTH dataset \textbf{(P)} pre-trained BERT model.}
    \label{tab:tesla2}
\end{table*}

On \Cref{tab:tesla2} we selected the passage from the ``Nikola\_Tesla" article\footnote{https://rajpurkar.github.io/SQuAD-explorer/explore/1.1/dev/Nikola\_Tesla.html} like Passage 2. This passage touches on the early life and background of Nikola Tesla. It mentions his birthdate, family background, and some of the influences and factors that contributed to his development and abilities. 

When analyzing the results for Question-3.1, DisGeM stands out for its ability to craft distractors closely linked to the answer ``Croatia", capturing the essence of the modern-day country of Tesla's birth. However, CDGP (F) seems to focus narrowly on the geographical proximity, offering choices like ``france" and ``hungary". While they are geographically related, they don't resonate with contextual clues like ``Austrian Empire", ``Smiljan", ``Serbian", and ``Serb family". Meanwhile, CDGP (P) offers distractors that closely align with the context, but critically includes ``Croatia", which is the actual answer, rendering the question invalid.

Turning to Question-3.2, DisGeM's distractors align well not only with the context and the flow but also with the answer ``his mother's genetics". In contrast, the distractors generated by both CDGP (F) and CDGP (P) are not as contextually fitting.

\subsection{Conclusion}

In summary, our qualitative analyses, supported by numerous examples, underline DisGeM's proficiency in generating contextually relevant and well-aligned distractors. DisGeM notably outperforms both CDGP (P) and CDGP (F), enhancing the quality of distractor generation, especially in scenarios where single-word distractor generation is inadequate. We highlighted several limitations of the CDGP framework, the previous SOTA, such as its strong reliance on fine-tuning, its tendency to produce distractors semantically close to the answer, and its inability to generate multi-word distractors. These insights reinforce the benefits of adopting the DisGeM framework, especially when high precision and relevance in distractor generation are paramount. Overall, DisGeM's predominant advantage stems from its capacity for multi-word generation.

\end{document}